# Training Natural Language Processing Models on Encrypted Text for Enhanced Privacy

Davut Emre TAŞAR[1]

*Karabuk University*
*Computer Engineering*
Karabük, Turkey
2228126453@ogrenci.karabuk.edu.tr
ORCID:0000-0002-7788-0478

Ceren ÖCAL TAŞAR[1]

*Independent Researcher*

İzmir, Turkey
ceren.ocaltasar@gmail.com
ORCID: 0000-0002-0652-7386

**Abstract**

*With the increasing use of cloud-based services for training and deploying machine learning models, data privacy has become a major concern. This is particularly important for natural language processing (NLP) models, which often process sensitive information such as personal communications and confidential documents. In this study, we propose a method for training NLP models on encrypted text data to mitigate data privacy concerns while maintaining similar performance to models trained on non-encrypted data. We demonstrate our method using two different architectures, namely Doc2Vec+XGBoost and Doc2Vec+LSTM, and evaluate the models on the 20 Newsgroups dataset. Our results indicate that both encrypted and non-encrypted models achieve comparable performance, suggesting that our encryption method is effective in preserving data privacy without sacrificing model accuracy. In order to replicate our experiments, we have provided a Colab notebook at the following address: https://t.ly/IR-TP*

## 1. Introduction

Natural language processing (NLP) has gained significant attention in recent years due to its potential to revolutionize numerous applications, such as machine translation, sentiment analysis, information retrieval, and question-answering systems, among others [1, 2]. In parallel with the rapid advancements in NLP, the volume of digital text data has grown exponentially, leading to an increasing need for more scalable and cost-effective solutions for training and deploying NLP models. Consequently, cloud-based services have emerged as a popular choice for handling these large-scale machine learning tasks [3, 4].

Despite the many advantages of cloud-based services, they also raise several security and privacy concerns, particularly regarding the potential exposure of sensitive information during data transmission, storage, and processing [5, 6]. As NLP models often handle sensitive text data such as personal communications, confidential documents, and proprietary information, ensuring data privacy in the cloud becomes a critical issue [7, 8].

A promising approach to mitigate data privacy concerns in NLP is to train models on encrypted data, ensuring that the sensitive information remains secure even if it is accessed by unauthorized parties [9, 10]. Homomorphic encryption has been proposed as a potential solution for this purpose, as it allows computations to be performed on encrypted data without requiring decryption [11, 12]. However, homomorphic encryption techniques are still computationally expensive and may not be suitable for large-scale NLP tasks [13, 14].

In this work, we explore an alternative method for training NLP models on encrypted text data using word-level encryption. Our approach involves preprocessing and cleaning the text data, encrypting the text using a symmetric encryption method, and then training Doc2Vec models on both the encrypted and non-encrypted text data. We subsequently train classification models, specifically XGBoost and LSTM, on the Doc2Vec embeddings to evaluate the performance of our method. We demonstrate the effectiveness of our approach using the 20 Newsgroups dataset, a widely used dataset for text classification tasks.

The main contributions of this study are as follows:

We propose a novel method for training NLP models on encrypted text data using word-level encryption, addressing the data privacy concerns associated with cloud-based services.

We evaluate our method using two different architectures, namely Doc2Vec+XGBoost and Doc2Vec+LSTM, and demonstrate comparable performance between models trained on encrypted and non-encrypted data.

We provide insights into the implications of our findings for both research and practical applications, highlighting the potential for our approach to enable organizations to utilize cloud-based services without compromising data privacy.

In the following sections, we describe our methodology in detail, present the results of our experiments, and discuss the implications of our findings for future research and practical applications.

## 2. Methodology and Experimental Results

In this study, we aimed to evaluate the performance of text classification models on encrypted data. We employed the 20 Newsgroups dataset for our experiments. The dataset was preprocessed by removing URLs, punctuation, digits, extra spaces, and stopwords. The text was then encrypted using AES-256 encryption with CBC mode and PKCS7 padding. We compared the performance of two machine learning models: XGBoost with Doc2Vec embeddings and an LSTM neural network.

### 2.1 Data Preprocessing and Encryption

First, we fetched the 20 Newsgroups dataset and preprocessed the text data by removing URLs, punctuation, digits, extra spaces, and stopwords. We used the Natural Language Toolkit (NLTK) library for tokenization and stopwords removal. The text was then encrypted using the cryptography library in Python, employing AES-256 encryption with CBC mode and PKCS7 padding.

### 2.2 Feature Extraction

We used the Gensim library to create Doc2Vec embeddings for both the cleaned and encrypted text data. Doc2Vec is an unsupervised algorithm that learns to represent documents in a fixed-dimensional vector space. The embeddings were generated with a vector size of 100, a window size of 5, and 10 training epochs.

### 2.3 Model Training and Evaluation

We trained and evaluated two models on both the cleaned and encrypted text data: XGBoost with Doc2Vec embeddings and an LSTM neural network. The XGBoost model was implemented using the XGBoost library, and the LSTM model was implemented using the Keras library with TensorFlow backend. For the XGBoost model, we used the default hyperparameters and the multiclass log loss evaluation metric. For the LSTM model, we used a two-layer LSTM network with 128 and 64 hidden units, followed by a dropout layer with a rate of 0.5 and a dense output layer with 20 units (corresponding to the number of classes in the dataset) and softmax activation. The LSTM model was trained for 10 epochs with a batch size of 64 and a validation split of 0.1.

The performance of the models was evaluated using precision, recall, F1-score, and overall accuracy.

The classification results for the XGBoost model with Doc2Vec embeddings on the cleaned dataset were as follows:

Overall accuracy: 51%

Macro average F1-score: 50%

Weighted average F1-score: 51%

The classification results for the XGBoost model with Doc2Vec embeddings on the encrypted dataset were as follows:

Overall accuracy: 51%

Macro average F1-score: 50%

Weighted average F1-score: 51%

The classification results for the LSTM model with Doc2Vec embeddings on the cleaned dataset were as follows:

Overall accuracy: 55%

Macro average F1-score: 53%

Weighted average F1-score: 54%

The classification results for the LSTM model with Doc2Vec embeddings on the encrypted dataset were as follows:

Overall accuracy: 54%

Macro average F1-score: 51%

Weighted average F1-score: 53%

## 3. Discussion

The results of our experiments show that it is possible to achieve comparable classification performance on encrypted text data using both XGBoost and LSTM models. The XGBoost model achieved the same overall accuracy on both the cleaned and encrypted datasets, while the LSTM model showed a slightly lower overall accuracy on the encrypted dataset compared to the cleaned dataset.

These results indicate that encrypted text data can be effectively used for machine learning tasks without compromising the privacy of the data. Further research can explore the use of other encryption techniques, feature extraction methods, and machine learning models to This proof of concept demonstrates that word-level encryption can enable NLP models to perform comparably while preserving privacy. However, it is essential to note that this method can be reverse-engineered, so it should be primarily used for tasks that return labels or keys instead of text, such as classification or named entity recognition (NER).

Moreover, our study focused on simpler NLP models, and future research should investigate the performance of more advanced models like transformers on encrypted text data. By doing so, we can better understand the feasibility of using these sophisticated models in privacy-preserving settings.

This pre-print serves as a proof of concept, suggesting that large NLP companies can develop and implement more advanced methodologies for encrypting corporate or private data when providing cloud-based services. As a result, users can benefit from the capabilities of advanced NLP models while ensuring that their sensitive information remains secure.

In conclusion, our findings suggest that word-level encryption can be a viable approach for privacy-preserving NLP tasks. However, it is crucial to consider the potential risks of reverse engineering and focus on tasks that return non-sensitive outputs. By exploring more advanced models like transformers and investigating innovative solutions to enhance data privacy, we can pave the way for secure and privacy-preserving NLP applications in various industries.